\documentclass[sigconf,anonymous=false]{acmart}

\definecolor{Gray}{gray}{0.9}
\usepackage{multirow}
\usepackage{color, colortbl}
\usepackage{xspace}
\usepackage{algorithm}
\usepackage{algorithmic}
\usepackage{float}
\usepackage{bm}
\newcommand{\shortsectionBf}[1]{\vspace{2pt}
\noindent {\bf #1}}
\newcommand{\sys}{{\textsc{\small{PersFL}}}\xspace}

\newcommand{\sysGD}{{\textsc{\small{PersFL-GD}}}\xspace}

\newcommand{\algrule}[1][.2pt]{\par\vskip.5\baselineskip\hrule height #1\par\vskip.5\baselineskip}

\usepackage{comment}
\usepackage{microtype}
\mathchardef\mhyphen="2D
\usepackage{wasysym}
\usepackage{stfloats}

\usepackage{tikz}
\DeclareRobustCommand*\circled[1]{\tikz[baseline=(char.base)]{ \node[shape=circle,draw,color=white,fill=black,inner sep=0.5pt] (char){#1};}}
\usepackage{subcaption}

\usepackage[scaled=.8]{beramono} 
\def\ie{{i.e.},~}

\DeclareMathOperator*{\argmin}{arg\,min}

\AtBeginDocument{%
  \providecommand\BibTeX{{%
    \normalfont B\kern-0.5em{\scshape i\kern-0.25em b}\kern-0.8em\TeX}}}

\begin{document}

\title{Unifying Distillation with Personalization in Federated Learning}


\author{Siddharth Divi}
\affiliation{%
  \institution{Purdue University}
  \streetaddress{State Street}
  \city{West Lafayette}
  \country{USA}}
\email{sdivi@purdue.edu}

\author{Habiba Farrukh}
\affiliation{%
  \institution{Purdue University}
  \city{West Lafayette}
  \country{USA}}
\email{hfarrukh@purdue.edu}

\author{Z Berkay Celik}
\affiliation{%
  \institution{Purdue University}
  \city{West Lafayette}
  \country{USA}}
\email{zcelik@purdue.edu}


\begin{abstract}
Federated learning (FL) is a decentralized privacy-preserving learning technique in which clients learn a joint collaborative model through a central aggregator without sharing their data. 
In this setting, all clients learn a single common predictor (\texttt{FedAvg}), which does not generalize well on each client's local data due to the statistical data heterogeneity among clients. 
In this paper, we address this problem with \sys, a discrete two-stage personalized learning algorithm. In the first stage,  \sys finds the optimal teacher model of each client during the FL training phase. In the second stage, \sys distills the useful knowledge from optimal teachers into each user's local model. 
The teacher model provides each client with some rich, high-level representation that a client can easily adapt to its local model, which overcomes the statistical heterogeneity present at different clients.
We evaluate \sys on CIFAR-10 and MNIST datasets using three data-splitting strategies to control the diversity between clients' data distributions.
We empirically show that \sys outperforms \texttt{FedAvg} and three state-of-the-art personalization methods, \texttt{pFedMe}, \texttt{Per-FedAvg} and \texttt{FedPer} on majority data-splits with minimal communication cost.
Further, we study the performance of \sys on different distillation objectives, how this performance is affected by the equitable notion of fairness among clients, and the number of required communication rounds.
\sys code is available at \url{https://tinyurl.com/hdh5zhxs} for public use and validation. 
%
\end{abstract}

\settopmatter{printacmref=false} 
\renewcommand\footnotetextcopyrightpermission[1]{} 

\maketitle
\pagestyle{plain} 

\section{Introduction}

\label{introduction}
Federated Learning (FL) is a distributed collaborative learning paradigm that does not require centralized data storage in a single location. Instead, a joint global predictor is learned jointly by a network of participating users~\cite{fedAvg}.
This paradigm is useful when the clients have private data that they cannot share with the participating entities due to privacy concerns. 
%
Recently, FL has found widespread applications in domains ranging from healthcare, finance to predictive keyboards. 

Federated Averaging (\texttt{FedAvg})~\cite{fedAvg} is an algorithm in which $\mathtt{n}$ users along with a central global aggregator participate together to learn a joint collaborative model. The data of each user does not leave their device. 
The users train the shared model on their local data and then share their model weights to the central aggregator. The central aggregator then aggregates (averages) the model updates from all the participating users and shares the new global shared model's updated weights. This process continues till convergence. In the end, each user obtains the same global model.

\shortsectionBf{Statistical Data Heterogeneity Problem.} 
FL faces different challenges such as expensive communication, systems heterogeneity, statistical heterogeneity, and privacy concerns~\cite{flSurvey_1}. 
Among these, statistical data heterogeneity has recently gained attraction, which means that clients' data are unbalanced and non-identical and independently distributed (non-IID).
Thus, the global model trained on clients' non-IID data restricts the global FL model from delivering good generalization on each client's local data. Each client gets a common model, irrespective of their data distribution.
For instance, consider the next word prediction engine that outputs what word comes next suggestions on a smartphone that enables users to express themselves faster.  
A common model learned collaboratively among clients fails to give each user useful suggestions, particularly when they have a unique way of expressing themselves in mobile applications such as in writing texts or emails. 
On the other hand, learning without client collaboration leads to a poor generalization of local clients due to a lack of data.
%
Personalized learning schemes proposed for FL aim to address this problem by finding a personalized model for each client that benefits from other clients' data while overcoming the statistical heterogeneity problem.
%

\shortsectionBf{Personalization Approaches.} 
There have been a few different approaches that proposed learning schemes through meta-learning, local fine-tuning,  multi-task learning,  model regularization, contextualization, and model interpolation to build personalized models.
For instance, \texttt{Per-FedAvg} uses Meta-Learning~\cite{metaFL_1, metaFL_2} to learn a common initialization point for all the users, which is then adapted to each user with a couple of steps of gradient descent. 
Another approach, \texttt{pFedMe}~\cite{pFedMe}, re-formulates the FL objective as a bi-level optimization problem and modifies the minimized loss function with the inclusion of a regularization term. 
Lastly, \texttt{FedPer}~\cite{fedPer} splits a deep neural network into base and personalization layers, where the base layers are learned collaboratively, and personalization layers are specific to each user.
However, some of these approaches incur high computational and algorithmic complexity. For instance, \texttt{pFedMe} and \texttt{Per-FedAvg} require a higher number of global communication rounds compared to \texttt{FedPer}.
Model Agnostic Meta-Learning based methods (\texttt{MAML})~\cite{maml} (used in \texttt{Per-FedAvg}) require the computation of the Hessian matrix, which significantly adds computational complexity to each client.
%

\shortsectionBf{Contributions.}
In this paper, we introduce \sys,  a new discrete two-stage personalization algorithm, which distills each client's optimal teacher model into each client's local model. 
In the first stage, each client participates in the FL training and stores the global model from the aggregator at each communication iteration. 
At the end of the FL training, clients measure each global model's validation error and set the model that gives the least error as an optimal teacher. 
Teacher models contain useful information unique to each client that can be readily adapted into the local models.
At the end of the first stage, each client obtains a separate teacher model and proceeds to the second stage.
In the second stage, each client distills the information from the optimal teacher model into their local model to learn a personalized model.
\sys controls the trade-off between optimal teacher and local model with temperature parameter that scales the class probability predictions from the teacher, and imitation parameter that balances how much a client imitates the teacher.
At the end of \sys algorithm, each client trains a local model based on their dataset and the useful knowledge extracted from other clients' datasets.

We empirically demonstrate the effectiveness of \sys using CIFAR-10 and MNIST datasets, which are widely used to evaluate personalized models' performance. 
We compare \sys with \texttt{FedAvg} and three recent approaches of personalization in FL, a transfer-learning based algorithm (\texttt{FedPer}), a bi-level optimization based algorithm (\texttt{pFedMe}), and a meta-learning based algorithm (\texttt{Per-FedAvg}).
To have a fair comparison, we use three different data-splitting strategies to control how each client's local dataset differs from other clients' data.

Our extensive experiments demonstrate that \sys outperforms \texttt{FedAvg}, and outperforms or yields comparable results with the \texttt{FedPer}, \texttt{pFedMe} and \texttt{Per-FedAvg} in local accuracy.
For example, compared to the best performing methods on CIFAR-10 data-splits, \sys improves the accuracy of users on average by $\mathtt{4.7\%}$, $\mathtt{0.6\%}$ and $\mathtt{3.9\%}$ over \texttt{FedPer}, \texttt{pFedMe}, and \texttt{FedPer}, respectively.
We perform additional experiments to characterize the equitable notion of fairness--the deviation among per-user accuracy, study the performance of variants of distillation objectives and investigate the number of communication rounds for convergence.
For instance, \sys reduces the deviation of per-user accuracy distributions on average by $\mathtt{1.5x}$ and $\mathtt{1.67x}$ compared to \texttt{Per-FedAvg} and \texttt{FedPer} algorithms on two different data-splits on MNIST.
For the number of global communication rounds, \sys takes $\mathtt{0.03x}$ and $\mathtt{0.5x}$ less communication rounds than \texttt{pFedMe} and \texttt{FedPer} on two different data-splits on CIFAR-10.
We show that these results challenge the existing objectives of personalized learning schemes and motivate new problems in personalization for the research community.

\section{Related Work}
Several prior works have explored techniques for personalization in FL instead of using a common model for all users. 
These works can be broadly grouped into the following categories based on the techniques adapted to improve clients' model performance, such as personalization via meta-learning, local fine-tuning,  multi-task learning,  model regularization, contextualization, and model interpolation.
Below, we review the recent core approaches.

\texttt{Per-FedAvg}~\cite{metaFL_2} uses \texttt{MAML} to learn an initialization point for each user, adapted to their local data after the training phase. In a closely related work, \texttt{Personalized FedAvg}~\cite{jiangetal}, a variant of Federated Averaging through Reptile algorithm, interprets \texttt{FedAvg} as a linear combination of a naive baseline and existing \texttt{MAML} methods.

\texttt{pFedMe}~\cite{pFedMe} formulates a bi-level optimization problem using the Moreau envelope as users' regularized loss function, thereby separating personalized model optimization from the global model learning.
\texttt{MOCHA}~\cite{fl_mtl} extends multi-task learning to the FL setting to overcome the ill-effects of statistical heterogeneity. Though this scheme leads to more user-specific solutions, it requires all the participating users to be available at all times during training. 

Adaptive Personalized Federated Learning (\texttt{APFL})~\cite{apfl} learns a personalized model for each user that is a mixture of the local and global models. The optimal mixing parameter, which controls the local and global models' ratio, is integrated into the learning problem. Another approach \texttt{FedPer}~\cite{fedPer} divides a deep neural network into the base and personalization layers to learn the base layers collaboratively and personalization layers specific to each user.

In a recent work~\cite{mansouretal}, three different approaches have been proposed with generalization guarantees, user clustering, data interpolation, and model interpolation. The first two approaches are not suitable for FL since they require meta-feature information from the clients, which raises privacy concerns. The third approach interpolates the local and global models and is closely related to the formulation of \texttt{APFL} method. 

Another work, \texttt{LotteryFL}~\cite{lotteryFL} adopts a Lottery Ticket Network through Lottery Ticket Hypothesis~\cite{lotteryTicketHypothesis} to learn personalized models for each user.
Lastly, \texttt{FedBE}~\cite{fedBE} integrates Bayesian learning into models to perform an aggregation of client uploaded model weights where the central aggregator creates a Bayesian model ensemble based on the client models. 

In contrast to the previous works, we use a modified formulation of the federated learning problem that incorporates distillation with each user's unique optimal teacher to train a per-user model based on their local dataset and other users' datasets.

\begin{algorithm*}[t!]
  \caption{\sys Algorithm}
  \label{alg:PersFL}
\begin{algorithmic}[1]
    \STATE \underline{\textbf{Stage-$\mathtt{1}$}}:\textbf{Finding the optimal teacher models} \\
    \STATE {\bfseries Notation:}\\
    $\mathtt{K}$: Clients indexed by $\mathtt{k}$, $\mathtt{E_{G}}$: Number of global aggregation rounds, $\mathtt{\mathcal{G}_{e}}$: Global \texttt{FedAvg} model during the global aggregation round $\mathtt{e}$, $\mathtt{O_{k}}$: Optimal teacher model for user $\mathtt{k}$, $\mathtt{l_{k'}}$: Loss of $\mathtt{O_{k}}$ on the user $\mathtt{k}$'s validation data (initialized to an arbitrarily large value), $\mathtt{x^{val}_{k}}$: Validation data of user $\mathtt{k}$.\\
    \FOR{global aggregation round $\mathtt{e=1}$ {\bfseries to} $\mathtt{E_{G}}$}
        \FOR{user $\mathtt{k=1}$ {\bfseries to} $\mathtt{K}$}
            \STATE $\mathtt{l_{k} \leftarrow \mathcal{L}_{cross}}(\sigma(\mathcal{G}_{e}(x^{val}_{k})), y^{val}_{k})$
            \IF{$\mathtt{l_{k} < l_{k'}}$}
                \STATE $\mathtt{l_{k'} \leftarrow l_{k}}$ 
                \STATE $\mathtt{O_{k} \leftarrow \mathcal{G}_{e}}$
            \ENDIF
        \ENDFOR
    \ENDFOR
    
    \algrule

    \STATE \underline{\textbf{Stage-$\mathtt{2}$}}: \textbf{Distilling from the optimal teacher models} \\
    \STATE {\bfseries Notation:}\\
    $\mathtt{K}$: Clients indexed by $\mathtt{k}$, $\mathtt{B}$: Local mini-batch size, $\mathtt{E_{L}}$: Number of local epochs, $\mathtt{T}$: Temperature parameter,
    $\mathtt{\lambda}$: Imitation parameter, $\mathtt{O_{k}}$: Optimal teacher model for user $\mathtt{k}$, $\mathtt{A_{k}}$: Local personalized model of user $\mathtt{k}$,
    $\mathtt{|T|_{c}}$: Search space of temperature values,
    \bm{$|\lambda|_{c}$}: Search space of imitation parameter values.\\
    
    {\bfseries Client Side:} \\
    \begin{itemize}
        \item For each client $\mathtt{k}$, initialize $\mathtt{A_{k}}$ with the weights of $\mathtt{O_{k}}$.\\
        
        \item Find the optimal values of $\mathtt{\lambda}$ and $\mathtt{T}$ by performing the following steps.\\
    \end{itemize}
    \FOR{user $\mathtt{k=1}$ {\bfseries to} $\mathtt{K}$}
        \STATE $\mathtt{B}$ $\leftarrow$ Obtain mini-batches of data from the user $\mathtt{k}$'s training data\\
      \FOR{local epoch $\mathtt{i=1}$ {\bfseries to} $\mathtt{E_{L}}$}
          \FOR{batch $\mathtt{b}$ $\in$ $\mathtt{B}$}
                \STATE $\mathtt{\mathcal{L}(A_{k},b) \leftarrow (1-\lambda)*hardLoss + (\lambda T^{2})*softLoss}$ 
                \STATE $\mathtt{hardLoss \leftarrow \mathcal{L}_{cross}(y_{b}, \sigma(A_{k}(b)))}$ 
                \STATE $\mathtt{softLoss \leftarrow \mathcal{KL}(s_{b}, \sigma(\frac{A_{k}(b)}{T}}))$ 
                \STATE $\mathtt{s_{b} \leftarrow \sigma(\frac{O_{k}(b)}{T})}$  
            \STATE $\mathtt{w \leftarrow w - \eta \nabla \mathcal{L}(A_{k}, b)}$
          \ENDFOR
      \ENDFOR
    \ENDFOR
\end{algorithmic}
\end{algorithm*}

\section{Preliminaries}
\label{preliminaries}
Model compression~\cite{modelCompression_caruana} or distillation~\cite{knowledgeDistillation_Hinton} are techniques to reduce the size and complexity of machine learning models. 
Distillation compresses a large complex model or an ensemble of models $\mathtt{f_{\text{large}}(x)}$ (teacher model) into a smaller and less complex model  $\mathtt{f_{\text{small}}(x)}$ (student model), which mimics the predictions of the complex model. 
There are scenarios in which the teacher model and ensemble models are too complicated from a computational perspective. Thus, distilling a large model into a simpler model makes it easier to run on limited computational resources such as on edge and mobile devices.
Remarkably, model distillation achieves model compression with no or minimal loss in accuracy.

Put in math, given a large model $\mathtt{f_{\text{large}}}$ that has been already learned, a small model $\mathtt{f_\text{small}}$ is learned by minimizing
%

\begin{multline}
 \mathtt{f_{s} = \argmin_{f_{small} \in F_{s}} \frac{1}{n} \sum^{n}_{i=1}[(1-\lambda) \mathcal{L}(\{(x_i,y_i)\}_{i=1}^n, f_{\text{small}})} + \\
 \mathtt{\lambda \mathcal{L}(\{x_i,s_i\}_{i=1}^n, f_{\text{small}})]}
 \label{eqn:eqn_1}
\end{multline}
where $\mathtt{f_{small}}$ is the candidate student model from the class capacity measure of the student model $\mathtt{F_{s}}$ (\ie hypothesis space of models for the student model ), $\mathtt{f_{s}}$ is the optimal student model learned, and $\mathtt{\lambda \in [0,1]}$ is the imitation parameter which controls how much $\mathtt{f_{small}}$ imitates $\mathtt{f_{large}}$ compared to directly learning from the data. 

In Equation \ref{eqn:eqn_1}, there are two datasets on which $\mathtt{f_{small}}$ is trained,  $\mathtt{{(x_{i}, y_{i})}_{i=1}^{n}}$ and $\mathtt{{(x_{i}, s_{i})}_{i=1}^{n}}$. 
$\mathtt{s_{i}}$ is called the soft label of the $\mathtt{f_{large}}$ model computed as $\mathtt{s_{i} = f_{large}(x_{i})/{T}}$ and $\mathtt{y_{i}}$ is the ground truth label. 
$\mathtt{T>0}$ is called the temperature parameter, which softens the teacher model's class probabilities. 
The softened class-probability predictions reveal dependencies among the labels that are otherwise hidden as either being extremely small or large numbers.

\section{Unifying Distillation with Personalization}
We introduce \sys, a new personalization algorithm, which unifies distillation with personalization to improve the generalization of each user's model accuracy instead of using the same global \texttt{FedAvg} model for each user. 
%

\subsection{\textsc{PersFL} Algorithm}
Our idea of distillation for personalization is inspired by the discrete phase local adaptation technique called the greedy local fine-tuning method~\cite{apfl}.
%
In this approach, a global \texttt{FedAvg} model is first learned during the training phase. During the subsequent adaptation phase, users perform several gradient descent steps to adapt the global \texttt{FedAvg} model's weights to users' local data distribution. 
However, a crucial question that needs to be answered is: \emph{why do all users adapt the same global model when the goal is personalization for each user?} 
This question inspires us to integrate a separate optimal teacher model into the local model of each user. 
To obtain the optimal teacher, each user iteratively validates the global \texttt{FedAvg} model on their local data during each aggregation round of the FL training phase. 
Subsequently, each user identifies the best global \texttt{FedAvg} model as an optimal teacher based on the accuracy.
%
Each user then incorporates the optimal teacher into their local model through distillation. 
Algorithm~\ref{alg:PersFL} details the steps of the \sys, which is a discrete two-stage algorithm.
%
%

\shortsectionBf{Stage-$\mathtt{1}$: Finding the optimal teacher models.}
In the first stage, each user participates in the training phase of FL and receives the current copy of the global model (\texttt{FedAvg}) in each round. In this step, each user locally stores the global model's current copy before updating the local version of the global model and sending it to the server for aggregation. After the termination of the FL training phase, each user finds the optimal teacher model $\mathtt{O_{k}}$ by minimizing:

\begin{equation}
    \mathtt{
        O_{k} = \argmin_{O_{k' \in |G|_{E}}} \mathcal{L}_{cross} (\sigma(O_{k'}(x^{val}_{k})), y^{val}_{k})
    }
    \label{eqn:eqn_3}    
\end{equation}

\noindent where $\mathtt{|G|_{E}}$ is the \texttt{FedAvg} model learned in each global aggregation round, $\mathtt{E}$ is the total number of global aggregation rounds of FL, $\mathtt{x^{val}_{k}}$ is the validation data of user $\mathtt{k}$ and $\mathtt{y^{val}_{k}}$ is the ground truth of the validation data of user $\mathtt{k}$. Thus, the optimal teacher model represents the global \texttt{FedAvg} model across the aggregation rounds that achieves the minimum loss on the validation data of user $\mathtt{k}$.



\shortsectionBf{Stage-$\mathtt{2}$: Distilling from the optimal teacher models.}
The second stage takes place locally for each user after the FL training phase, independent of FL (\ie no client collaboration). 
We call this stage local adaptation. We first initialize the personalized model, $\mathtt{A_{k}}$ with the weights of $\mathtt{O_{k}}$ for each user $\mathtt{k}$. Following this, each user distills the information from the optimal teacher ($\mathtt{O_k}$) into their local model to learn $\mathtt{A_{k}}$.
Specifically, each user computes $\mathtt{A_{k}}$ by distilling hard-loss and soft-loss by minimizing:

{\small{
\begin{multline}
        \mathtt{
            A_{k} = \underset{\underset{A_{k'} \in |A_{k}|_{C}}{\lambda' \in |\lambda|_{C}, T' \in |T|_{C}}}{\argmin}
            } 
            \overbrace{\mathtt{(1 - \lambda') (\mathcal{L}_{cross}(\sigma(A_{k'}(x^{train}_{k})), y^{train}_{k}))}}^\textrm{hard-loss}
        \\
           + 
           \overbrace{\mathtt{(\lambda' T'^{2}) \times \mathcal{KL}(\sigma(\frac{A_{k'}(x^{train}_{k})}{T'}),\sigma(\frac{O_{k'}(x^{train}_{k})}{T'}))}}^\textrm{soft-loss}
\label{eqn:eqn_4}
\end{multline}
}}

The hard-loss refers to the loss of the student model ($\mathtt{A_{k}}$) on the hard-labels ($\mathtt{y_{i}}$), whereas soft-loss refers to the loss of the student model on the soft labels ($\mathtt{s_{i}}$). 
Soft labels are the teacher model's ($\mathtt{O_{k}}$) scaled predictions.
$\mathtt{|\lambda|_{C}}$ refers to the complexity of the search space of the imitation parameter, $\mathtt{|T|_{C}}$ denotes the search space complexity of the temperature parameter, and $\mathtt{|A_{k}|_{C}}$ refers to the hypothesis space of the personalized models for $\mathtt{k}$. 
Here, each user performs a grid-based search to find the optimal values for distillation parameters 
temperature ($\mathtt{T}$) and imitation parameter ($\mathtt{\lambda}$) while  simultaneously learning the personalized model.

We use Kullback-Leibler (KL) divergence between the output of the teacher model ($\mathtt{O_{k}}$) and the student model ($\mathtt{A_{k}}$). 
Enforcing the logits of $\mathtt{O_{k}}$ and $\mathtt{A_{k}}$ to be similar yields a regularizing effect, which in turn improves the generalization ability of $\mathtt{A_{k}}$. 
We additionally multiply the soft-loss by $\mathtt{T^{2}}$ since the gradients of the term $\mathtt{\sigma({A_{k}(b)}/{T})}$ scale as $\mathtt{{1}/{T^{2}}}$. 
This ensures that the relative contributions of the hard-labels ($\mathtt{y_{i}}$) and the soft-labels ($\mathtt{s_{i}}$) are roughly unchanged if the $\mathtt{T}$ value changes. 
%
%

At the end of the \textbf{Stage-$\mathtt{2}$}, each user learns a personalized model optimized for their local data distribution by distilling the useful knowledge from other users into the personalized model $\mathtt{A_{k}}$ through the optimal teacher $\mathtt{O_{k}}$ learned in \textbf{Stage-$\mathtt{1}$}.

\subsection{Why Does \textsc{PersFL} Work?}
We introduce the learning under privileged information (LUPI) paradigm~\cite{vladimirPrivilegedInformation} and show that \sys reduces to Generalized Distillation~\cite{generalizedDistillation_david}, an instance of the LUPI paradigm.  
Vapnik's LUPI paradigm assumes that feature-label pairs $\mathtt{(x_{i}, y_{i})}$, and additional information $\mathtt{x_{i}^{\star}}$ about $\mathtt{(x_{i}, y_{i})}$ are available at training time and $\mathtt{x_{i}^{\star}}$ is not available at test time. Here, $\mathtt{x_{i}^{\star}}$ is called the privileged information.
For example,  consider the problem of identifying cancerous biopsy images.
$\mathtt{x}$ is the biopsy image of a patient in pixel space.
An oncologist may describe the biopsy image relevant to cancer in a specialized language space different than the pixel space. 
The descriptions are called privileged information ($\mathtt{x^{\star}}$), which contain useful information to classify the biopsy images, however this information is not available at test time.

Generalized distillation develops an objective to learn from multiple data representations as follows.
First, it learns a teacher model $\mathtt{f_{t}}$ on the feature-target set $\mathtt{\{x^\star_i, y_i\}_{i=1}^n}$. 
Second, it computes teacher soft labels $\mathtt{s_i = f_{t}(x^\star_i)/T}$ using a temperature parameter $\mathtt{T > 0}$.
Lastly, it learns a student model $\mathtt{f_{s}}$ from $\mathtt{\{x_i, y_i\}_{i=1}^n}$, $\mathtt{\{x_i, s_i\}_{i=1}^n}$.

In \sys, the optimal teacher model $\mathtt{O_{k}}$ of each user $\mathtt{k}$ is analogous to the privileged information $\mathtt{x^{\star}}$. 
Since each user's data distributions are not exactly the same, each user's optimal teacher model is unique and identified with the best performing \texttt{FedAvg} model on the user $\mathtt{i}$'s validation data during the FL communication rounds. 
A user can obtain the intricate patterns from other users' large amounts of data through the weights of $\mathtt{O_{k}}$ as privileged information, only available to each user during FL training.  

Once each user identifies the teacher model, the soft-labels of the teacher model ($\mathtt{s_{i}}$) are computed on the data of each user. 
\sys learns the student model from the teacher through distillation by choosing optimal values of the parameters, $\mathtt{\lambda}$ and $\mathtt{T}$. 
We argue that distilling information from the teacher model to the student model overcomes the \emph{catastrophic forgetting} problem~\cite{catastrophicForgetting}, which is the tendency of a model to forget the information learned in the previously trained tasks when it is trained on new tasks.
\sys mitigates this problem by first initializing the student model with the teacher model's weights and then distilling the teacher model's information to the student model.
\begin{figure}[t!]
\begin{center}
\centerline{\includegraphics[width=\columnwidth]{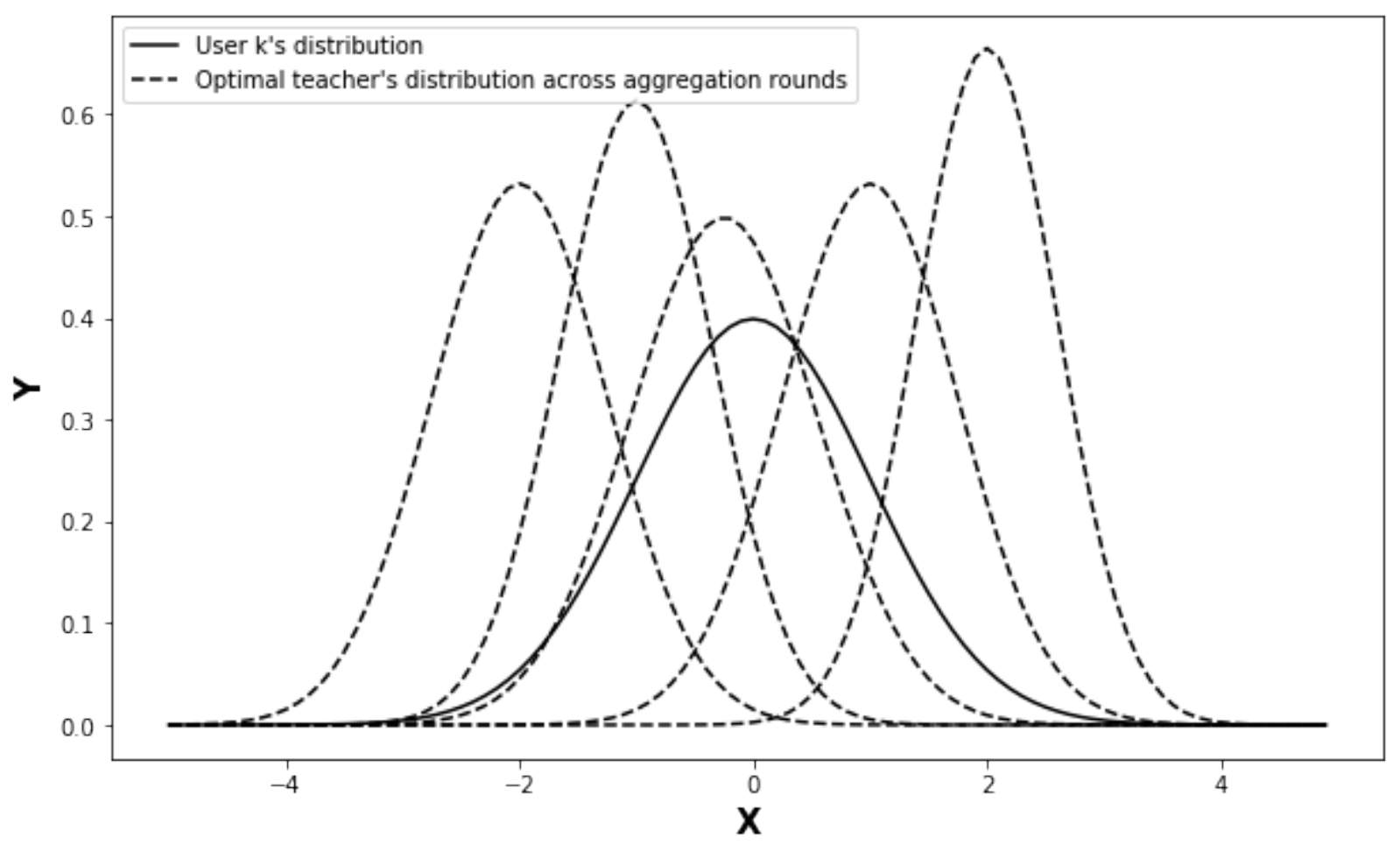}}
\caption{Users $\mathtt{k}$'s distribution vs. global \texttt{FedAvg} model's distribution across the global aggregation rounds. 
} 
\label{fig:fig_1}
\end{center}
\end{figure}

Figure~\ref{fig:fig_1} shows the training data distribution of a user $\mathtt{k}$ (solid line) and the distribution of the \texttt{FedAvg} model in the global communication rounds (dashed lines).
The global model's distribution comes close to approximating the user's distribution and starts moving away from it. 
This divergence of distributions is caused due to the non-IID data distribution across users. 
\sys's teacher model is unique to each user, and there is maximal overlap with their data distribution, 
which helps address the statistical heterogeneity problem and mitigate its negative transfer effect.
Our extensive experiments in Section~\ref{sec:evaluation} validate this hypothesis.

\subsection{Convergence and Complexity Analysis}
\shortsectionBf{Convergence of \textsc{PersFL}.} Convergence of an algorithm is defined as the algorithm's ability to converge to the global optimum, which is defined as the region in the loss landscape with the lowest possible loss (global minima). 
Equation~\ref{eqn:flObjective} represents the core objective of the FL setting. Here, $\mathtt{f_{i}(.)}$ denotes the expected loss over the data distribution of user $\mathtt{i}$ when there are $\mathtt{N}$ users, and $\mathtt{w}$ refers to the weights of the \texttt{FedAvg} model being learned.
\begin{equation}
    \mathtt{
        \underset{{w \in \mathcal{R}^{d}}}{min}\Bigg{\{}f(w) := \frac{1}{N}\sum_{i=1}^{N}f_{i}(w)\Bigg{\}}
    }
    \label{eqn:flObjective}    
\end{equation}

Unlike other personalized FL methods such as \texttt{pFedMe} and \texttt{FedPer}, the convergence analysis for \sys is not needed since \sys does not modify the FL core objective.
To detail, \sys is a two-stage discrete algorithm in which the users join the FL training in the first stage to learn a separate optimal teacher model. Each user then independently distills the optimal teacher to their local dataset. Therefore, the convergence of \sys is the same as the convergence of the \texttt{FedAvg} algorithm~\cite{fedAvgConvergence}.

\shortsectionBf{Complexity of \textsc{PersFL}.} We analyze the complexity of \sys in terms of the number of epochs executed locally at each user during the training phase.
The worst-case complexity of \sys when all users are computationally involved in every round is given by $\mathtt{\mathcal{O}(E_{G}E_{L} + |T|_{C}|\lambda|_{C}E)}$. $\mathtt{E_{G}}$ is the number of global communication rounds, $\mathtt{E_{L}}$ is the number of local epochs, $\mathtt{|T|_{C}}$ and $\mathtt{|\lambda|_{C}}$ are the class complexity measures of the search space and $\mathtt{E}$ is the number of epochs that we set to distill information from the optimal teacher model into the personalized model. 

\section{Experiments}
\label{sec:evaluation}
We evaluate the performance of \sys using two datasets, each with three data-splits, and compare its results with \texttt{FedAvg} and three recent personalization approaches, \texttt{FedPer}, \texttt{pFedMe},  and \texttt{Per$\mhyphen$FedAvg}. 
%
Table~\ref{tab:relatedWork_table1} describes the compared approaches with \sys, including the datasets used in their evaluation, data-splits (detailed below), and the algorithms that they compare with their techniques.

We conduct all the experiments with $10$ users. 
We make three assumptions in line with the assumptions made in compared approaches. 
First, we assume that all users are active during the entire training phase to speed up the model convergence. 
Second, the data of each user does not change between the global aggregations. 
Lastly, the hyper-parameters, batch-size ($\mathtt{B}$), and local epochs ($\mathtt{E}$), are invariant among the participant users. 
We conduct all experiments with a $\mathtt{60\%}$-$\mathtt{20\%}$-$\mathtt{20\%}$ train-validation-test split on an NVIDIA Tesla T4 GPU with $\mathtt{16GB}$ memory.

\shortsectionBf{Datasets and Datasplits.}
We evaluate \sys on the CIFAR-10 and MNIST datasets. These datasets are widely used in FL training and are also used by the compared methods.
To have a fair comparison, we use three different data-splits for both the CIFAR-10 and MNIST. MNIST is a dataset of $\mathtt{28 \times 28}$ images of handwritten digits from $\mathtt{0}$-$\mathtt{9}$ consisting of $\mathtt{10}$ labels and $\mathtt{70,000}$ instances. CIFAR-10 is a dataset of $\mathtt{32 \times 32}$ color images with $\mathtt{10}$ classes and $\mathtt{60,000}$ instances. 

In data-split 1 (\textbf{DS-1}), each user has the same total number of samples but may have different classes and a different number of samples per class.
The statistical heterogeneity is varied by controlling the parameter $\mathtt{k}$, which controls the number of overlapping classes between each user. 
For example, $\mathtt{k=4}$ corresponds to a highly non-identical data partition, whereas $\mathtt{k=10}$ corresponds to a highly identical data partition across the participating users. In our experiments, we set $\mathtt{k}$ to 4 to have non-IID data across users. 

%
\begin{table}[t!]
\centering
\setlength\tabcolsep{1.5pt} 
\def\arraystretch{1.2} 
\resizebox{\columnwidth}{!}
{\begin{tabular}{|l|c|c|c|c|c|}\hline
\multicolumn{1}{|c|}{\textbf{\#}}
&\multicolumn{1}{|c|}{\textbf{Method}} 
&\multicolumn{1}{|c|}{\textbf{Datasets}} &\multicolumn{1}{|c|}{\textbf{Comparison}} &\multicolumn{1}{|c|}{\textbf{Datasplit}} \\\hline\hline
$1$ & \textbf{Per-FedAvg}~\cite{metaFL_2}  & MNIST, CIFAR10 & \texttt{FedAvg} & Custom  \\\hline
$2$ &  \textbf{pFedMe}~\cite{pFedMe}  & MNIST, Synthetic dataset & \texttt{FedAvg}, \texttt{Per-FedAvg} & DS-3 \\\hline
$3$ &  \textbf{FedPer}~\cite{fedPer} & FLICKR-AES, CIFAR-10, CIFAR-100 & \texttt{FedAvg} & DS-1 \\\hline
\end{tabular}}
\caption{The details of the personalized federated learning methods compared with \sys.} 
\label{tab:relatedWork_table1}
\end{table}
In \textbf{DS-2}, all users have samples from all classes, but the number of samples per class they have is different, and hence the total number of samples per user is also different across users. 
%
In order to simulate a non-IID distribution, we assign samples from each class to the users using a Dirichlet distribution with $\mathtt{\alpha = 0.9}$, following the previous work~\cite{nIID_dist}. Each class is parameterized by a vector $\mathtt{q}$ where $\mathtt{q \geq 0, i \in [1, N]}$, where $\mathtt{q}$ is sampled from a Dirichlet distribution with parameters $\mathtt{\alpha}$ and $\mathtt{p}$. 
The parameter $\mathtt{p}$ is the prior class distribution over the classes, and $\mathtt{\alpha}$ is the concentration parameter that controls the data similarity among the users. 
If $\mathtt{\alpha \rightarrow \infty}$, all users have an identical distribution to the prior. If $\mathtt{\alpha \rightarrow 0}$, each user only has samples from one class randomly chosen. 

For \textbf{DS-3}, each user has two of the ten class-labels. 
Additionally, the total number of samples per user is different, \ie all the users do not have the same number of total samples. 
The samples assigned to users are drawn from a log-normal distribution with the parameters $\mathtt{\mu = 0}$ and $\mathtt{\sigma = 2}$.
A variable $\mathtt{u}$ has a log-normal distribution if $\mathtt{log(u)}$ is normally distributed. 
The probability density function for the log-normal distribution is computed as:

\begin{equation}
    \mathtt{p(u) = \frac{1}{\sigma u \sqrt{(2\pi)}} e^{-(\frac{(ln(u) - \mu)^2}{2\sigma^{2}})}}
    \label{eqn:eqn_2}    
\end{equation}

These parameters correspond to the underlying normal distribution from which we draw the samples.

\begin{table*}[th!]\centering
\def\arraystretch{1.04}
\setlength\tabcolsep{8pt}
\resizebox{\textwidth}{!}
{\begin{tabular}{l|rrr|rrr|rrr|rrr|rrrr}\toprule
\textbf{} &\multicolumn{3}{c|}{\textbf{FedAvg}} &\multicolumn{3}{c|}{\textbf{\sys}} &\multicolumn{3}{c|}{\textbf{FedPer}} &\multicolumn{3}{c|}{\textbf{pFedMe}} &\multicolumn{3}{c}{\textbf{Per-FedAvg}} \\
\cmidrule{1-16}
\textbf{Users} &\textbf{DS-1} &\textbf{DS-2} &\textbf{DS-3} &\textbf{DS-1} &\textbf{DS-2} &\textbf{DS-3} &\textbf{DS-1} &\textbf{DS-2} &\textbf{DS-3} &\textbf{DS-1} &\textbf{DS-2} &\textbf{DS-3} &\textbf{DS-1} &\textbf{DS-2} &\textbf{DS-3} \\
\midrule
\rowcolor{Gray}
\textbf{User 0} &43.6 &50.8 &48.2 &85.5 &61.3 &94.5 &83.2 &57.2 &93.1 &74.3 &61.7 &94.2 &69.2 &58.2 &92.5 \\
\textbf{User 1} &50.9 &45.3 &40.8 &78.2 &56.9 &79.9 &74.5 &51.4 &77 &64.1 &57.3 &79.2 &65 &56.1 &73.7 \\
\rowcolor{Gray}
\textbf{User 2} &44.5 &49.4 &31.2 &82.2 &57.3 &68.9 &78.4 &53.2 &64.7 &69.6 &57.3 &64.4 &67.9 &57.2 &64.6 \\
\textbf{User 3} &51.3 &46.5 &31.5 &82.1 &60.1 &82.5 &77.9 &55.4 &77.5 &69.4 &58.9 &72.5 &67.2 &58.8 &77 \\
\rowcolor{Gray}
\textbf{User 4} &45.3 &50.8 &49.4 &79.4 &59.1 &82.5 &76.1 &54.4 &78.6 &67.2 &59.5 &80 &65.8 &59.4 &82.5 \\
\textbf{User 5} &44.2 &50.7 &47.8 &77.1 &61.9 &79.9 &72.1 &57.6 &76.9 &62.1 &60.6 &77.5 &62.7 &59.3 &77.9 \\
\rowcolor{Gray}
\textbf{User 6} &35.8 &46.5 &56.8 &75.6 &58.9 &90.3 &70.9 &53.3 &88.5 &59.9 &58.6 &88.3 &58.2 &57.7 &89.1 \\
\textbf{User 7} &37.9 &49.5 &58.1 &79.7 &61.2 &87.6 &75.6 &56.9 &84.6 &65.8 &60.2 &84 &64.3 &58 &83.7 \\
\rowcolor{Gray}
\textbf{User 8} &47.7 &48.7 &49 &87.7 &60 &76.7 &84.4 &57.5 &73.5 &75.5 &59.6 &66.9 &72.5 &55.8 &64.1 \\
\textbf{User 9} &48.6 &49.1 &53.5 &91 &58.8 &80.3 &88.5 &54.3 &77.9 &81.7 &58.3 &73.8 &76.6 &55.3 &72.7 \\
\midrule
\textbf{Avg. Acc.} &\textbf{45} &\textbf{48.7} &\textbf{46.6} &\textbf{81.9} &\textbf{59.6} &\textbf{82.3} &\textbf{78.2} &\textbf{55.1} &\textbf{79.2} &\textbf{69} &\textbf{59.2} &\textbf{78.1} &\textbf{66.9} &\textbf{57.6} &\textbf{77.8} \\
\midrule
\textbf{Std Dev} &\textbf{5.1} &\textbf{2} &\textbf{9.4} &\textbf{4.9} &\textbf{1.7} &\textbf{7.2} &\textbf{5.6} &\textbf{2.1} &\textbf{7.9} &\textbf{6.7} &\textbf{1.4} &\textbf{9.2} &\textbf{5.1} &\textbf{1.5} &\textbf{9.5} \\
\bottomrule
\end{tabular}}
\caption{Accuracy per user for the different personalization algorithms on the different data-splits on the CIFAR-10 dataset.}
\label{table:table1}
\end{table*}

\begin{table*}[th!]\centering
\setlength\tabcolsep{8pt}
\def\arraystretch{1.03}
\resizebox{\textwidth}{!}
{\begin{tabular}{l|rrr|rrr|rrr|rrr|rrrr}\toprule
\textbf{} &\multicolumn{3}{c|}{\textbf{FedAvg}} &\multicolumn{3}{c|}{\textbf{\sys}} &\multicolumn{3}{c|}{\textbf{FedPer}} &\multicolumn{3}{c|}{\textbf{pFedMe}} &\multicolumn{3}{c}{\textbf{Per-FedAvg}} \\
\cmidrule{1-16}
\textbf{Users} &\textbf{DS-1} &\textbf{DS-2} &\textbf{DS-3} &\textbf{DS-1} &\textbf{DS-2} &\textbf{DS-3} &\textbf{DS-1} &\textbf{DS-2} &\textbf{DS-3} &\textbf{DS-1} &\textbf{DS-2} &\textbf{DS-3} &\textbf{DS-1} &\textbf{DS-2} &\textbf{DS-3} \\
\midrule
\rowcolor{Gray}
\textbf{User 0} &91.2 &84.2 &98.3 &98.3 &88.6 &99.6 &98.2 &84.2 &99.9 &94.8 &88.2 &99 &97.3 &87.5 &98.6 \\
\textbf{User 1} &92.2 &85.2 &97.9 &98.8 &86.3 &99.3 &98.8 &85.2 &99.3 &95.8 &86.1 &98.3 &99 &86.1 &97.6 \\
\rowcolor{Gray}
\textbf{User 2} &88 &83.5 &96.2 &99.5 &86 &98.7 &98.3 &83.5 &98.2 &93.2 &85.4 &97.5 &99.7 &86.2 &96.8 \\
\textbf{User 3} &93.2 &84.4 &94.8 &98.2 &87.9 &99.6 &97.8 &84.4 &99.6 &94.3 &86.7 &97.6 &98.7 &87.5 &96 \\
\rowcolor{Gray}
\textbf{User 4} &92.3 &86.6 &96.3 &98.7 &88.1 &99.6 &98.2 &86.6 &99.6 &93.7 &87.4 &98.1 &99 &88 &97.1 \\
\textbf{User 5} &92.2 &84.5 &97.3 &98.3 &86.4 &99.2 &97.8 &84.5 &99.1 &94.2 &85.3 &98.5 &98.8 &86.8 &97.7 \\
\rowcolor{Gray}
\textbf{User 6} &93.8 &87 &97.7 &98.7 &88.8 &99.8 &98.5 &87 &99.9 &95 &88 &98.8 &99.2 &87.9 &98.2 \\
\textbf{User 7} &91.5 &86.8 &95.7 &98.7 &88.7 &99.6 &98.5 &86.8 &99.4 &95.8 &87.7 &98 &99.3 &88.2 &96.9 \\
\rowcolor{Gray}
\textbf{User 8} &94.3 &85.8 &94.3 &98.3 &89.1 &99.2 &98.3 &85.8 &98.9 &94.5 &87.9 &97.8 &98.8 &88.5 &96.5 \\
\textbf{User 9} &91.3 &86.7 &95.9 &98.5 &89.7 &99.7 &98.5 &86.7 &99.6 &96.2 &88.6 &98.3 &99.2 &89.1 &97.4 \\
\midrule
\textbf{Avg Acc} &\textbf{92} &\textbf{85.5} &\textbf{96.4} &\textbf{98.6} &\textbf{88} &\textbf{99.4} &\textbf{98.3} &\textbf{85.5} &\textbf{99.4} &\textbf{94.8} &\textbf{87.1} &\textbf{98.2} &\textbf{98.9} &\textbf{87.6} &\textbf{97.3} \\
\midrule
\textbf{Std Dev} &\textbf{1.7} &\textbf{1.3} &\textbf{1.3} &\textbf{0.4} &\textbf{1.3} &\textbf{0.3} &\textbf{0.3} &\textbf{1.3} &\textbf{0.5} &\textbf{1} &\textbf{1.2} &\textbf{0.5} &\textbf{0.6} &\textbf{1} &\textbf{0.8} \\
\bottomrule
\end{tabular}}
\caption{Accuracy per user for the different personalization algorithms on the different data-splits on the MNIST dataset.}
\label{table:table2}
\end{table*}

\shortsectionBf{Model Architecture.}
For the CIFAR-10 dataset, we use a CNN-based model with two 2-D convolutional layers separated by a MaxPool layer between them and followed by three fully connected (FC) layers.  
The fully connected layers have $400$, $120$, and $84$ hidden neurons. We use ReLu activations after each layer except the last FC layer.  
For the MNIST dataset, we use a two-layer deep neural network (DNN) with $100$ hidden-layer neurons. We use a ReLu activation on the hidden layer. 
The output layer has $10$ nodes with a softmax function to get the class probabilities. The architectures we use for both datasets are similar to those in \texttt{pFedMe} for conformity.

\subsection{Degree of Personalization across Users} 
To study the degree (extent) of personalization of \sys and to compare it with the other personalization approaches, we average experiments of CIFAR-10 and MNIST datasets over $5$ experimental runs for each data-split. 
Table~\ref{table:table1} and Table~\ref{table:table2} show the per-user accuracy of \texttt{Fed-Avg}, \sys, \texttt{FedPer}, \texttt{pFedMe}, and \texttt{Per-FedAvg} on different datasplits of CIFAR-10 and MNIST. 
Below, we present the performance of \sys with the \texttt{FedAvg} model and the best performing algorithm on the average accuracy across users. 

Our analysis of CIFAR-10 in Table~\ref{table:table1} shows that \sys performs better than other personalization techniques across all data-splits.
In comparison to the \texttt{FedAvg} model, \sys leads to a percentage increase of $\mathtt{82\%}$, $\mathtt{22.3\%}$, and $\mathtt{76.6\%}$ on \textbf{DS-1}, \textbf{DS-2}, and \textbf{DS-3}. 
%
For the compared approaches, for all data splits, \sys improves the on average accuracy by $\mathtt{4.7\%}$, $\mathtt{0.6\%}$ and $\mathtt{3.9\%}$ across users compared to \texttt{FedPer}, \texttt{pFedMe}, and \texttt{FedPer} respectively.
%
The absolute improvement of \sys over other techniques on \textbf{DS-1}, \textbf{DS-2} and \textbf{DS-3} is $\mathtt{3.7\%}$, $\mathtt{0.4\%}$ and $\mathtt{3.1\%}$, compared to \texttt{FedPer}, \texttt{pFedMe}, and \texttt{FedPer}.
%

The results of our experiments on MNIST in Table~\ref{table:table2} show that, in comparison to the \texttt{FedAvg} model, \sys improves the accuracy on average by $\mathtt{7.1\%}$, $\mathtt{2.9\%}$, and $\mathtt{3.1\%}$ on \textbf{DS-1}, \textbf{DS-2}, and \textbf{DS-3}.
\sys performs similar to the other approaches. 
In \textbf{DS-1}, the best-performing approach \texttt{Per-FedAvg} gives $\mathtt{98.9\%}$ accuracy, which performs slightly better than $\mathtt{98.6\%}$ accuracy of \sys.
In \textbf{DS-2}, \sys gives a $\mathtt{0.4\%}$ increase in accuracy across users compared to \texttt{Per-FedAvg}.  
In \textbf{DS-3}, both \sys and \texttt{FedPer} gives $\mathtt{99.4\%}$ accuracy.

\begin{table*}[!t]
    \begin{minipage}{.48\textwidth}
        \centering
        \setlength\tabcolsep{1.5pt}
        \resizebox{\columnwidth}{!}{%
        \begin{tabular}{lrrrrrrrrrrrrr}\toprule
        \textbf{} &\multicolumn{4}{c}{\textbf{DS-1}} &\multicolumn{4}{c}{\textbf{DS-2}} &\multicolumn{4}{c}{\textbf{DS-3}} \\\cmidrule{1-13}
        \textbf{Users} &\multicolumn{2}{c}{\textbf{FedAvg}} &\multicolumn{2}{c}{\textbf{Opt Teacher}} &\multicolumn{2}{c}{\textbf{FedAvg}} &\multicolumn{2}{c}{\textbf{Opt Teacher}} &\multicolumn{2}{c}{\textbf{FedAvg}} &\multicolumn{2}{c}{\textbf{Opt Teacher}} \\\midrule
        \rowcolor{Gray}
        \textbf{User 0} &\multicolumn{2}{c}{84.5} &\multicolumn{2}{c}{85.5} &\multicolumn{2}{c}{59.5} &\multicolumn{2}{c}{61.3} &\multicolumn{2}{c}{94.6} &\multicolumn{2}{c}{94.5} \\
        \textbf{User 1} &\multicolumn{2}{c}{77.6} &\multicolumn{2}{c}{78.2} &\multicolumn{2}{c}{55.9} &\multicolumn{2}{c}{56.9} &\multicolumn{2}{c}{78.7} &\multicolumn{2}{c}{79.9} \\
        \rowcolor{Gray}
        \textbf{User 2} &\multicolumn{2}{c}{81.2} &\multicolumn{2}{c}{82.2} &\multicolumn{2}{c}{55.7} &\multicolumn{2}{c}{57.3} &\multicolumn{2}{c}{68.2} &\multicolumn{2}{c}{68.9} \\
        \textbf{User 3} &\multicolumn{2}{c}{81.4} &\multicolumn{2}{c}{82.1} &\multicolumn{2}{c}{58.6} &\multicolumn{2}{c}{60.1} &\multicolumn{2}{c}{81} &\multicolumn{2}{c}{82.5} \\
        \rowcolor{Gray}
        \textbf{User 4} &\multicolumn{2}{c}{78.7} &\multicolumn{2}{c}{79.4} &\multicolumn{2}{c}{58.1} &\multicolumn{2}{c}{59.1} &\multicolumn{2}{c}{82} &\multicolumn{2}{c}{82.5} \\
        \textbf{User 5} &\multicolumn{2}{c}{75.2} &\multicolumn{2}{c}{77.1} &\multicolumn{2}{c}{60.9} &\multicolumn{2}{c}{61.9} &\multicolumn{2}{c}{79.6} &\multicolumn{2}{c}{79.9} \\
        \rowcolor{Gray}
        \textbf{User 6} &\multicolumn{2}{c}{74.4} &\multicolumn{2}{c}{75.6} &\multicolumn{2}{c}{56.6} &\multicolumn{2}{c}{58.9} &\multicolumn{2}{c}{90.1} &\multicolumn{2}{c}{90.3} \\
        \textbf{User 7} &\multicolumn{2}{c}{79.4} &\multicolumn{2}{c}{79.7} &\multicolumn{2}{c}{59.6} &\multicolumn{2}{c}{61.2} &\multicolumn{2}{c}{86.9} &\multicolumn{2}{c}{87.6} \\
        \rowcolor{Gray}
        \textbf{User 8} &\multicolumn{2}{c}{86.6} &\multicolumn{2}{c}{87.7} &\multicolumn{2}{c}{58.4} &\multicolumn{2}{c}{60} &\multicolumn{2}{c}{75.8} &\multicolumn{2}{c}{76.7} \\
        \textbf{User 9} &\multicolumn{2}{c}{90.4} &\multicolumn{2}{c}{91} &\multicolumn{2}{c}{56.3} &\multicolumn{2}{c}{58.8} &\multicolumn{2}{c}{79.5} &\multicolumn{2}{c}{80.3} \\
        \midrule
        \textbf{Avg Acc} &\multicolumn{2}{c}{\textbf{80.9}} &\multicolumn{2}{c}{\textbf{81.9}} &\multicolumn{2}{c}{\textbf{58}} &\multicolumn{2}{c}{\textbf{59.6}} &\multicolumn{2}{c}{\textbf{81.6}} &\multicolumn{2}{c}{\textbf{82.3}} \\
        \midrule
        \textbf{Std Dev} &\multicolumn{2}{c}{\textbf{5}} &\multicolumn{2}{c}{\textbf{4.9}} &\multicolumn{2}{c}{\textbf{1.8}} &\multicolumn{2}{c}{\textbf{1.7}} &\multicolumn{2}{c}{\textbf{7.5}} &\multicolumn{2}{c}{\textbf{7.2}} \\
        \bottomrule
        \end{tabular}
    }    
     \subcaption{CIFAR-10}
    \label{table:initBasedExp}
    \end{minipage}%
     \hspace{0.5cm}
    \begin{minipage}{.48\textwidth}
      \centering
     \setlength\tabcolsep{1.5pt}
        \resizebox{\columnwidth}{!}{%
        \begin{tabular}{lrrrrrrrrrrrrr}\toprule
        \textbf{} &\multicolumn{4}{c}{\textbf{DS-1}} &\multicolumn{4}{c}{\textbf{DS-2}} &\multicolumn{4}{c}{\textbf{DS-3}} \\\cmidrule{1-13}
        \textbf{Users} &\multicolumn{2}{c}{\textbf{FedAvg}} &\multicolumn{2}{c}{\textbf{Opt Teacher}} &\multicolumn{2}{c}{\textbf{FedAvg}} &\multicolumn{2}{c}{\textbf{Opt Teacher}} &\multicolumn{2}{c}{\textbf{FedAvg}} &\multicolumn{2}{c}{\textbf{Opt Teacher}} \\\midrule
        \rowcolor{Gray}
        \textbf{User 0} &\multicolumn{2}{c}{98.3} &\multicolumn{2}{c}{98.3} &\multicolumn{2}{c}{89} &\multicolumn{2}{c}{88.6} &\multicolumn{2}{c}{99.6} &\multicolumn{2}{c}{99.6} \\
        \textbf{User 1} &\multicolumn{2}{c}{98.8} &\multicolumn{2}{c}{98.8} &\multicolumn{2}{c}{86.3} &\multicolumn{2}{c}{86.3} &\multicolumn{2}{c}{99.3} &\multicolumn{2}{c}{99.3} \\
        \rowcolor{Gray}
        \textbf{User 2} &\multicolumn{2}{c}{99.3} &\multicolumn{2}{c}{99.5} &\multicolumn{2}{c}{86.3} &\multicolumn{2}{c}{86} &\multicolumn{2}{c}{98.7} &\multicolumn{2}{c}{98.7} \\
        \textbf{User 3} &\multicolumn{2}{c}{98.2} &\multicolumn{2}{c}{98.2} &\multicolumn{2}{c}{87.9} &\multicolumn{2}{c}{87.9} &\multicolumn{2}{c}{99.6} &\multicolumn{2}{c}{99.6} \\
        \rowcolor{Gray}
        \textbf{User 4} &\multicolumn{2}{c}{98.7} &\multicolumn{2}{c}{98.7} &\multicolumn{2}{c}{88.1} &\multicolumn{2}{c}{88.1} &\multicolumn{2}{c}{99.6} &\multicolumn{2}{c}{99.6} \\
        \textbf{User 5} &\multicolumn{2}{c}{98.3} &\multicolumn{2}{c}{98.3} &\multicolumn{2}{c}{86.7} &\multicolumn{2}{c}{86.4} &\multicolumn{2}{c}{99.2} &\multicolumn{2}{c}{99.2} \\
        \rowcolor{Gray}
        \textbf{User 6} &\multicolumn{2}{c}{98.7} &\multicolumn{2}{c}{98.7} &\multicolumn{2}{c}{89} &\multicolumn{2}{c}{88.8} &\multicolumn{2}{c}{99.8} &\multicolumn{2}{c}{99.8} \\
        \textbf{User 7} &\multicolumn{2}{c}{98.7} &\multicolumn{2}{c}{98.7} &\multicolumn{2}{c}{89} &\multicolumn{2}{c}{88.7} &\multicolumn{2}{c}{99.6} &\multicolumn{2}{c}{99.6} \\
        \rowcolor{Gray}
        \textbf{User 8} &\multicolumn{2}{c}{98.3} &\multicolumn{2}{c}{98.3} &\multicolumn{2}{c}{89} &\multicolumn{2}{c}{89.1} &\multicolumn{2}{c}{99.2} &\multicolumn{2}{c}{99.2} \\
        \textbf{User 9} &\multicolumn{2}{c}{98.5} &\multicolumn{2}{c}{98.5} &\multicolumn{2}{c}{89.6} &\multicolumn{2}{c}{89.7} &\multicolumn{2}{c}{99.7} &\multicolumn{2}{c}{99.7} \\
        \midrule
        \textbf{Avg Acc} &\multicolumn{2}{c}{\textbf{98.6}} &\multicolumn{2}{c}{\textbf{98.6}} &\multicolumn{2}{c}{\textbf{88.1}} &\multicolumn{2}{c}{\textbf{88}} &\multicolumn{2}{c}{\textbf{99.4}} &\multicolumn{2}{c}{\textbf{99.4}} \\
        \midrule
        \textbf{Std Dev} &\multicolumn{2}{c}{\textbf{0.3}} &\multicolumn{2}{c}{\textbf{0.4}} &\multicolumn{2}{c}{\textbf{1.2}} &\multicolumn{2}{c}{\textbf{1.3}} &\multicolumn{2}{c}{\textbf{0.3}} &\multicolumn{2}{c}{\textbf{0.3}} \\
        \bottomrule
        \end{tabular}
}
\subcaption{MNIST}
\label{table:initBasedExp_MNIST}
\end{minipage}%
 \caption{Average Accuracy across all the users for personalized models initialized with the weights of FedAvg model vs. Optimal Teacher model for each user on (a) CIFAR-10 and (b) MNIST datasets. 
 }
\end{table*}


\subsection{Equitable Notion of Fairness among Users}
To understand the distribution of the personalized models' performance across users, we compute the standard deviation (\texttt{SD}) of the per-user accuracy. 
From an equitable notion of fairness, this helps us understand how fair the personalization improvements are across users~\cite{qFFL}. 
For two personalization techniques $\mathtt{t}$ and $\mathtt{t'}$, the performance distribution among $\mathtt{K}$ users represented by $\mathtt{\{F_{1}(t), \ldots, F_{K}(t)\}}$ is more fair (uniform) under technique $\mathtt{t}$ than $\mathtt{t'}$ if the following holds: 
\begin{equation}
    \textbf{$\mathtt{Var}$}\mathtt{(F_{1}(t), ... F_{K}(t))} < \textbf{$\mathtt{Var}$}\mathtt{(F_{1}(t'), ... F_{K}(t'))}
\end{equation}

We compare \sys with the best performing method in terms of average accuracy.
In CIFAR-10 experiments (Table~\ref{table:table1}), \sys yields a \texttt{SD} of $\mathtt{4.9}$ among the average accuracy of users on \textbf{DS-1}, which is the least deviation compared to the other approaches.
This yields a reduction of $\mathtt{1.14x}$ in \texttt{SD} compared to \texttt{FedPer}.
For \textbf{DS-2}, \texttt{pFedMe} yields an \texttt{SD} of $\mathtt{1.4}$, slightly better than $\mathtt{1.7}$ \texttt{SD} of \sys.  
For \textbf{DS-3}, \sys achieves a reduction of $\mathtt{1.10x}$ in \texttt{SD} compared to \texttt{FedPer}.

For MNIST (Table~\ref{table:table2}), we observe similar results compared to the experiments on CIFAR-10.
For \textbf{DS-1}, although \texttt{Per-FedAvg} with an average $\mathtt{98.9\%}$ accuracy is better than \sys's $\mathtt{98.6\%}$ accuracy, \sys leads to a $\mathtt{1.5x}$ reduction in \texttt{SD} compared to it. 
For \textbf{DS-2}, we find that the \texttt{SD} of \texttt{Per-FedAvg} at $\mathtt{1.0}$, is lower than the \texttt{SD} of \sys at $\mathtt{1.3}$. On \textbf{DS-3}, the reduction factor for \sys in \texttt{SD} stands at $\mathtt{1.67x}$ compared to \texttt{FedPer}.
Overall, \sys often leads to a reduction in the SD of the average accuracy across users, thus leading to a more uniform (fair) distribution than the compared methods.

 
%




\subsection{Impact of Optimal Teachers on Accuracy}
%

We claim that each user's optimal teacher model ($\mathtt{O_{k}}$) has maximal overlap with their local data distribution. 
To understand the impact of optimal teachers on the degree of personalization of the personalized models, we conduct the following experiment. 
We compare two different ways of choosing the teacher model, \ie global \texttt{FedAvg} and optimal teacher models used in \textbf{Stage-$\mathtt{2}$} of \sys as the teacher for distillation. 
We observe that choosing the optimal teacher model as the teacher and performing distillation leads to more personalized solutions and a lower standard deviation in the per-user accuracy.

Table~\ref{table:initBasedExp} and Table~\ref{table:initBasedExp_MNIST} show the results of using \texttt{FedAvg} and optimal teacher for CIFAR-10 and MNIST datasets.
We make the following observations in terms of the average per-user accuracy. 
On \textbf{DS-1}, initializing with the optimal teacher model leads to a percentage increase of $\mathtt{1.2\%}$ and an absolute increase of $\mathtt{1\%}$. 
On \textbf{DS-2}, the optimal teacher initialization leads to a percentage increase of $\mathtt{2.7\%}$ and an absolute increase of $\mathtt{1.6\%}$. 
In the case of \textbf{DS-3}, there is a $\mathtt{0.8\%}$ percentage and $\mathtt{0.7\%}$ absolute increase attributed to initialization with optimal teachers. 
%
Additionally, we observe that \sys yields slightly lower standard deviations of per-user accuracy.
We conduct the same experiments on the MNIST dataset and present the results in Table~\ref{table:initBasedExp_MNIST}. 
The average accuracy across users and the standard deviations of the average accuracy are very similar across all the methods.
We argue that this variation can be attributed to the lack of inter-class variations in the MNIST dataset.

\begin{figure}[t!]
\begin{center}
\centerline{\includegraphics[width=1\columnwidth]{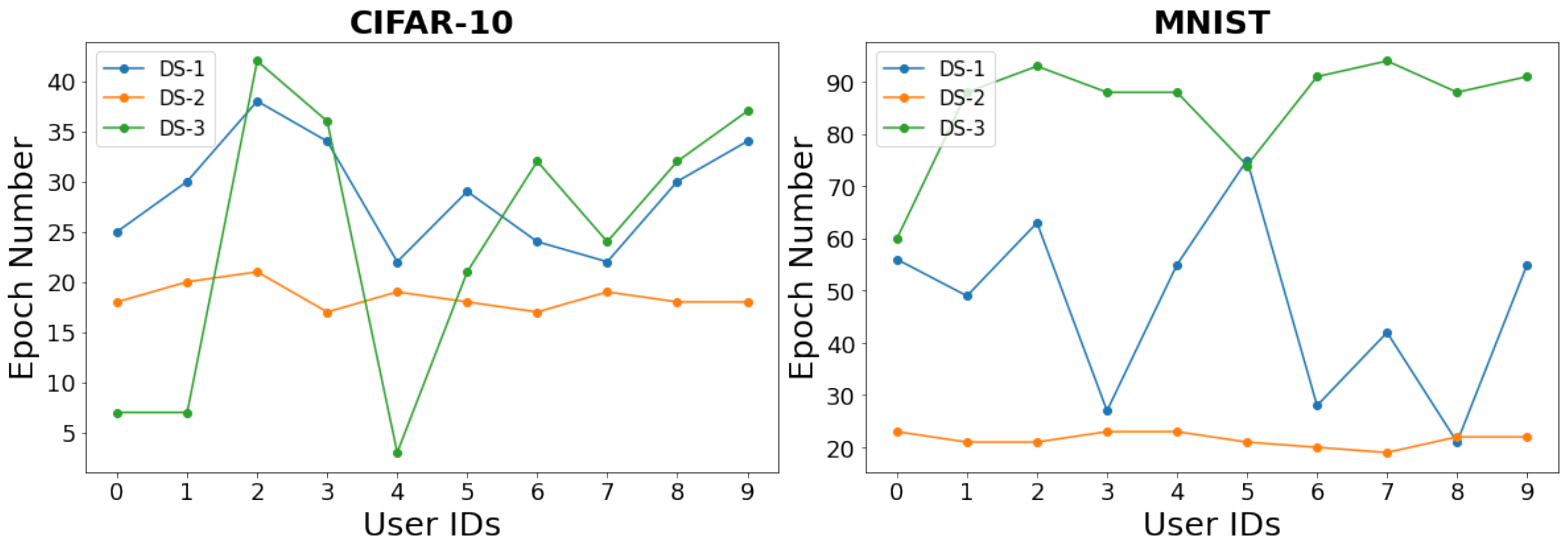}}
\caption{Average epoch numbers when users' optimal teacher models are selected across data-splits.}
\label{fig:optTeacherEpochs}
\end{center}
\end{figure}

Figure~\ref{fig:optTeacherEpochs} shows the averaged epoch numbers across the different experimental runs for both CIFAR-10 and MNIST, across all data-splits. 
The users choose their optimal teacher model after participating in the global communication rounds. 
We observe that the averaged epoch numbers in which the optimal teacher models are chosen are predominantly different. This observation validates our hypothesis that each user has a unique optimal teacher model, and all users should not use the same teacher model. 
Turning to Figure~\ref{fig:fig_1}, the dashed curve closest to the solid curve corresponds to the optimal teacher model for a particular user $\mathtt{k}$. 
The averaged epoch number across all experimental runs at which this optimal teacher model is chosen is shown in Figure \ref{fig:optTeacherEpochs}.

\begin{table}[!t]\centering
\def\arraystretch{1.1}
\resizebox{\columnwidth}{!}{%
\begin{tabular}{lrrrrrrrrrrrrrrrrrrrrrrrrr}\toprule
\textbf{} &\multicolumn{12}{c}{\textbf{CIFAR-10}} &\multicolumn{12}{c}{\textbf{MNIST}} \\\cmidrule{2-25}
\textbf{} &\multicolumn{4}{c}{\textbf{DS-1}} &\multicolumn{4}{c}{\textbf{DS-2}} &\multicolumn{4}{c}{\textbf{DS-3}} &\multicolumn{4}{c}{\textbf{DS-1}} &\multicolumn{4}{c}{\textbf{DS-2}} &\multicolumn{4}{c}{\textbf{DS-3}} \\\cmidrule{2-25}
\textbf{Users} &\multicolumn{2}{c}{\textbf{T}} &\multicolumn{2}{c}{\textbf{$\mathtt{\lambda}$}} &\multicolumn{2}{c}{\textbf{T}} &\multicolumn{2}{c}{\textbf{$\mathtt{\lambda}$}} &\multicolumn{2}{c}{\textbf{T}} &\multicolumn{2}{c}{\textbf{$\mathtt{\lambda}$}} &\multicolumn{2}{c}{\textbf{T}} &\multicolumn{2}{c}{\textbf{$\mathtt{\lambda}$}} &\multicolumn{2}{c}{\textbf{T}} &\multicolumn{2}{c}{\textbf{$\mathtt{\lambda}$}} &\multicolumn{2}{c}{\textbf{T}} &\multicolumn{2}{c}{\textbf{$\mathtt{\lambda}$}} \\\midrule
\rowcolor{Gray}
\textbf{User 0} &\multicolumn{2}{c}{12.2} &\multicolumn{2}{c}{0.45} &\multicolumn{2}{c}{8.2} &\multicolumn{2}{c}{0.6} &\multicolumn{2}{c}{6.6} &\multicolumn{2}{c}{0.25} &\multicolumn{2}{c}{1} &\multicolumn{2}{c}{0} &\multicolumn{2}{c}{11.4} &\multicolumn{2}{c}{0.45} &\multicolumn{2}{c}{1} &\multicolumn{2}{c}{0} \\
\textbf{User 1} &\multicolumn{2}{c}{8.2} &\multicolumn{2}{c}{0.4} &\multicolumn{2}{c}{25} &\multicolumn{2}{c}{0.7} &\multicolumn{2}{c}{12.2} &\multicolumn{2}{c}{0.5} &\multicolumn{2}{c}{1} &\multicolumn{2}{c}{0} &\multicolumn{2}{c}{5.8} &\multicolumn{2}{c}{0.6} &\multicolumn{2}{c}{1} &\multicolumn{2}{c}{0} \\
\rowcolor{Gray}
\textbf{User 2} &\multicolumn{2}{c}{7.4} &\multicolumn{2}{c}{0.45} &\multicolumn{2}{c}{12.2} &\multicolumn{2}{c}{0.65} &\multicolumn{2}{c}{2.6} &\multicolumn{2}{c}{0.3} &\multicolumn{2}{c}{1.8} &\multicolumn{2}{c}{0.1} &\multicolumn{2}{c}{15.4} &\multicolumn{2}{c}{0.4} &\multicolumn{2}{c}{1} &\multicolumn{2}{c}{0} \\
\textbf{User 3} &\multicolumn{2}{c}{17} &\multicolumn{2}{c}{0.3} &\multicolumn{2}{c}{21} &\multicolumn{2}{c}{0.6} &\multicolumn{2}{c}{17} &\multicolumn{2}{c}{0.35} &\multicolumn{2}{c}{1} &\multicolumn{2}{c}{0} &\multicolumn{2}{c}{10.6} &\multicolumn{2}{c}{0.45} &\multicolumn{2}{c}{1} &\multicolumn{2}{c}{0} \\
\rowcolor{Gray}
\textbf{User 4} &\multicolumn{2}{c}{17} &\multicolumn{2}{c}{0.5} &\multicolumn{2}{c}{11.4} &\multicolumn{2}{c}{0.75} &\multicolumn{2}{c}{11.4} &\multicolumn{2}{c}{0.15} &\multicolumn{2}{c}{1} &\multicolumn{2}{c}{0.05} &\multicolumn{2}{c}{11.4} &\multicolumn{2}{c}{0.6} &\multicolumn{2}{c}{1} &\multicolumn{2}{c}{0} \\
\textbf{User 5} &\multicolumn{2}{c}{12.2} &\multicolumn{2}{c}{0.55} &\multicolumn{2}{c}{16.2} &\multicolumn{2}{c}{0.6} &\multicolumn{2}{c}{20.2} &\multicolumn{2}{c}{0.35} &\multicolumn{2}{c}{1} &\multicolumn{2}{c}{0.05} &\multicolumn{2}{c}{1.8} &\multicolumn{2}{c}{0.65} &\multicolumn{2}{c}{5.8} &\multicolumn{2}{c}{0.05} \\
\rowcolor{Gray}
\textbf{User 6} &\multicolumn{2}{c}{20.2} &\multicolumn{2}{c}{0.6} &\multicolumn{2}{c}{17} &\multicolumn{2}{c}{0.7} &\multicolumn{2}{c}{7.4} &\multicolumn{2}{c}{0.2} &\multicolumn{2}{c}{1} &\multicolumn{2}{c}{0.05} &\multicolumn{2}{c}{11.4} &\multicolumn{2}{c}{0.4} &\multicolumn{2}{c}{1} &\multicolumn{2}{c}{0} \\
\textbf{User 7} &\multicolumn{2}{c}{12.2} &\multicolumn{2}{c}{0.4} &\multicolumn{2}{c}{2.6} &\multicolumn{2}{c}{0.65} &\multicolumn{2}{c}{4.2} &\multicolumn{2}{c}{0.1} &\multicolumn{2}{c}{1} &\multicolumn{2}{c}{0} &\multicolumn{2}{c}{11.4} &\multicolumn{2}{c}{0.45} &\multicolumn{2}{c}{1} &\multicolumn{2}{c}{0} \\
\rowcolor{Gray}
\textbf{User 8} &\multicolumn{2}{c}{12.2} &\multicolumn{2}{c}{0.5} &\multicolumn{2}{c}{21} &\multicolumn{2}{c}{0.75} &\multicolumn{2}{c}{2.6} &\multicolumn{2}{c}{0.35} &\multicolumn{2}{c}{1} &\multicolumn{2}{c}{0.1} &\multicolumn{2}{c}{3.4} &\multicolumn{2}{c}{0.6} &\multicolumn{2}{c}{1} &\multicolumn{2}{c}{0} \\
\textbf{User 9} &\multicolumn{2}{c}{20.2} &\multicolumn{2}{c}{0.3} &\multicolumn{2}{c}{17} &\multicolumn{2}{c}{0.75} &\multicolumn{2}{c}{8.2} &\multicolumn{2}{c}{0.6} &\multicolumn{2}{c}{1} &\multicolumn{2}{c}{0} &\multicolumn{2}{c}{6.6} &\multicolumn{2}{c}{0.5} &\multicolumn{2}{c}{1} &\multicolumn{2}{c}{0} \\
\bottomrule
\end{tabular}}
\caption{Average optimal values of $\mathtt{\lambda}$ and $\mathtt{T}$ for each user across different experimental runs on CIFAR-10 and MNIST.}
\label{table:optParams}
\end{table}

\subsection{Optimal Parameter Selection}
\label{optParamSelection}
We conduct experiments to understand distillation's effectiveness by investigating the values of the optimal distillation parameters chosen by each user's student model.
Table~\ref{table:optParams} shows the average values of distillation parameters, imitation ($\mathtt{\lambda}$) and temperature ($\mathtt{T}$), averaged out over the five experimental runs on CIFAR-10 and MNIST datasets. 
For CIFAR-10, we observe that the $\mathtt{\lambda}$ parameter values are non-zero across users and data-splits. 
This observation means that distillation is effective, and the teacher model's knowledge is useful in learning the student model. 
In contrast, in the case of MNIST, we observe that $\mathtt{\lambda}$ parameter values are mostly very close to $\mathtt{0}$ in the case of \textbf{DS-1} and \textbf{DS-3}. 
We believe that this is because of the nature of the MNIST dataset \ie due to the lack of inter-class variations. 
%
We show a detailed analysis of the variation of the classification accuracy on the test-set across users depending on the values of the distillation parameters $\mathtt{\lambda}$ and $\mathtt{T}$ in the Appendix~\ref{appendix-1}.




\subsection{Variants of \textsc{PersFL}}
The distillation objective of \sys in Equation~\ref{eqn:eqn_4} can be implemented with different objective functions defining how to distill the information from the optimal teacher model into the user's local model.
To demonstrate the generalization of \sys to different distillation objectives, we perform the distillation with a different loss function and compare it with Equation~\ref{eqn:eqn_4}.

We define the following objective function between the soft labels and the student's soft predictions:

\vspace{-7pt}{\small{
\begin{multline}
        \mathtt{
            A_{k} = \underset{\underset{A_{k'} \in |A_{k}|_{C}}{\lambda' \in |\lambda|_{C}, T' \in |T|_{C}}}{\argmin}
            } 
            \overbrace{\mathtt{(1 - \lambda') (\mathcal{L}_{cross}(\sigma(A_{k'}(x^{train}_{k})), y^{train}_{k}))}}^\textrm{hard-loss}
        \\
          + 
          \overbrace{\mathtt{(\lambda' T'^{2}) \times \mathcal{L}_{cross}(\sigma(\frac{A_{k'}(x^{train}_{k})}{T'}),\sigma(\frac{O_{k'}(x^{train}_{k})}{T'}))}}^\textrm{soft-loss}
\label{eqn:altDistillation}
\end{multline}
}}\vspace{-7pt}

The soft predictions of a student are defined as the predictions of the student model, which are scaled by the temperature parameter, $\mathtt{\sigma(A_k'(x^{train}_k)/T')}$. 
Equation~\ref{eqn:altDistillation} is different from the original formulation of \sys in Equation~\ref{eqn:eqn_4}. 
Here we do not use KL-divergence to enforce the similarity of logits between the teacher and student model. 
In contrast, we compute the cross-entropy between the models.
We refer to this variant of \sys to \texttt{\sysGD}. We note that other distillation methods can be easily integrated into \sys. 

\begin{table}[!htp]\centering
\setlength\tabcolsep{1.5pt}
\resizebox{\columnwidth}{!}{%
\begin{tabular}{lrrrrrrrrrrrrr}\toprule
&\multicolumn{6}{c}{\textbf{CIFAR-10}} &\multicolumn{6}{c}{\textbf{MNIST}} \\\cmidrule{2-13}
\textbf{} &\multicolumn{2}{c}{\textbf{DS-1}} &\multicolumn{2}{c}{\textbf{DS-2}} &\multicolumn{2}{c}{\textbf{DS-3}} &\multicolumn{2}{c}{\textbf{DS-1}} &\multicolumn{2}{c}{\textbf{DS-2}} &\multicolumn{2}{c}{\textbf{DS-3}} \\\cmidrule{2-13}
\textbf{Users} &\textbf{\circled{1}} &\circled{2} &\circled{1} &\circled{2} &\circled{1} & \circled{2} &\circled{1} &\circled{2} &\circled{1} &\circled{2} &\circled{1} &\circled{2} \\\midrule
\rowcolor{Gray}
\textbf{User 0} &85.5 &85.5 &61.3 &60 &94.5 &94.1 &98.3 &98.3 &88.6 &88.1 &99.6 &99.8 \\
\textbf{User 1} &78.2 &77.8 &56.9 &56.4 &79.9 &78.8 &98.8 &99 &86.3 &85.3 &99.3 &99.4 \\
\rowcolor{Gray}
\textbf{User 2} &82.2 &81.6 &57.3 &56.2 &68.9 &68.2 &99.5 &99.2 &86 &85.5 &98.7 &98.7 \\
\textbf{User 3} &82.1 &81.5 &60.1 &58.4 &82.5 &82 &98.2 &98.2 &87.9 &87.2 &99.6 &99.8 \\
\rowcolor{Gray}
\textbf{User 4} &79.4 &79.2 &59.1 &58.2 &82.5 &82.4 &98.7 &98.5 &88.1 &87.5 &99.6 &99.7 \\
\textbf{User 5} &77.1 &76.3 &61.9 &61 &79.9 &78.8 &98.3 &98.7 &86.4 &85.6 &99.2 &99.3 \\
\rowcolor{Gray}
\textbf{User 6} &75.6 &75.5 &58.9 &57.9 &90.3 &90.1 &98.7 &98.5 &88.8 &88.3 &99.8 &99.9 \\
\textbf{User 7} &79.7 &79.5 &61.2 &58.8 &87.6 &87 &98.7 &99 &88.7 &88.2 &99.6 &99.7 \\
\rowcolor{Gray}
\textbf{User 8} &87.7 &87.1 &60 &58.4 &76.7 &75.6 &98.3 &98.8 &89.1 &88.4 &99.2 &99.3 \\
\textbf{User 9} &91 &90.5 &58.8 &56.3 &80.3 &79.5 &98.5 &98.8 &89.7 &88.7 &99.7 &99.8 \\
\midrule
\textbf{Avg Acc} &\textbf{81.9} &\textbf{81.5} &\textbf{59.6} &\textbf{58.2} &\textbf{82.3} &\textbf{81.7} &\textbf{98.6} &\textbf{98.7} &\textbf{88} &\textbf{87.3} &\textbf{99.4} &\textbf{99.5} \\
\midrule
\textbf{Std Dev} &\textbf{4.9} &\textbf{4.9} &\textbf{1.7} &\textbf{1.6} &\textbf{7.2} &\textbf{7.4} &\textbf{0.4} &\textbf{0.3} &\textbf{1.3} &\textbf{1.3} &\textbf{0.3} &\textbf{0.4} \\
\bottomrule
\end{tabular}}
\caption{Performance comparison of variants of \sys. \circled{1} is for  \sys, and \circled{2} is for \sysGD objective functions.}
\label{tab:KDvsGD}
\end{table}

Table~\ref{tab:KDvsGD} shows the results of both variants of \sys across all data-splits on both CIFAR-10 and MNIST datasets. 
We make the following three observations. 
First, in MNIST, both variants of \sys on \textbf{DS-1} and \textbf{DS-3} have a very similar performance in terms of average accuracy across all users.
In \textbf{DS-2} \sys has an absolute performance improvement of $\mathtt{0.7\%}$ compared to \sysGD. 
Second, the performance difference is relatively more in CIFAR-10 experiments.
\sys has an absolute improvement of $\mathtt{1.4\%}$ on \textbf{DS-2} of CIFAR-10 and an improvement of $\mathtt{0.6\%}$ on \textbf{DS-3}. 
In all cases, \sys performs better than or at least equal to \sysGD.
Lastly, in terms of the standard deviations between the per-user accuracy, both \sys and \sysGD yield very similar performance.  
%

\begin{table}[t!]
\centering
\setlength\tabcolsep{9pt}
\resizebox{\columnwidth}{!}{%
\begin{tabular}{l|c|c|c|c|c|c|}
\cline{2-7}
& \multicolumn{3}{c|}{\textbf{CIFAR-10}} & \multicolumn{3}{c|}{\textbf{MNIST}} \\ \hline
\multicolumn{1}{|l|}{\textbf{Method}} & \textbf{DS-1} & \textbf{DS-2} & \textbf{DS-3} & \textbf{DS-1} & \textbf{DS-2} & \textbf{DS-3} \\ \hline \hline
\multicolumn{1}{|l|}{\textbf{FedAvg}} &50 &25 &100 &100 &50 &100 \\
\multicolumn{1}{|l|}{\textbf{\sys}}&50 &25 &50 &100 &25 &100 \\
\multicolumn{1}{|l|}{\textbf{FedPer}} &50 &25 &100 &100 &50 &100 \\
\multicolumn{1}{|l|}{\textbf{pFedMe}} &800 &1000 &800 &800 &800 &800 \\
\multicolumn{1}{|l|}{\textbf{Per-FedAvg}} &800 &800 &1000 &800 &800 &800 \\
\bottomrule
\end{tabular}
}
\caption{Comparison of \# of global communication rounds}
\label{table:commRounds}
\end{table}

\subsection{Global Communication Rounds}
%

In this set of experiments, we compare the number of global communication rounds taken by each personalization algorithm. 
Table~\ref{table:commRounds} shows the number of global communication rounds for each algorithm.
Below, for each dataset's data-splits, we compare the number of global communication rounds required by \sys with the best performing method.
%
We observe that, in CIFAR-10 for \textbf{DS-1}, \sys takes $50$ communication rounds similar to \texttt{FedPer}. 
For \textbf{DS-2}, \sys takes $\mathtt{0.03x}$ less communication rounds than \texttt{pFedMe}, and for \textbf{DS-3}, \sys takes $\mathtt{0.5x}$ less communication rounds than \texttt{FedPer}.
In MNIST experiments, we observe that for \textbf{DS-1} \sys needs $\mathtt{0.13x}$ less communication rounds required by \texttt{Per-FedAvg}. 
In \textbf{DS-2}, \sys requires $\mathtt{0.03x}$ less communication rounds compared to \texttt{Per-FedAvg}, and \sys needs the same number of communication rounds as \texttt{FedPer} in \textbf{DS-3}.

\section{Discussion and Limitations}

\shortsectionBf{Personalization for New Participants.} 
\sys can easily adapt new users participating in the FL framework to learn personalized models.
To detail,  consider that a new user $\mathtt{k'}$ joins the framework after the training phase of \sys in search of a personalized solution. 
Since each user chooses their optimal teacher model at the end of the \sys training phase, the \texttt{FedAvg} model for user $\mathtt{k'}$ may serve as the teacher model \ie{$\mathtt{O_{k'}}$} (assuming that only the \texttt{FedAvg} model in the final global aggregation round is stored).
If the central aggregator stores the \texttt{FedAvg} model across the global aggregation rounds, user $\mathtt{k'}$ can then choose $\mathtt{O_{k'}}$ to be the most optimal \texttt{FedAvg} model across all the global aggregation rounds that has the lowest error on the validation data of user $\mathtt{k'}$. 
Following this, user $\mathtt{k'}$ can then easily perform a search over $\mathtt{\lambda}$ and $\mathtt{T}$ to find the optimal values within the search space according to Equation~\ref{eqn:eqn_4}. 
Subsequently, the user can perform distillation with $\mathtt{O_{k'}}$ to learn a personalized model $\mathtt{A_{k'}}$.
However, the global \texttt{FedAvg} model learned over the aggregation rounds may not be the most optimal teacher model for the new participants when the number of new participants joining the framework increases.
In future work, we will study the trade-off between the number of new participants and the accuracy of personalized models with respect to the shift in their data distribution.


\shortsectionBf{Dual Optimization of \sys.} 
Our work raises some new important questions, such as how to unify distillation with personalization in FL such that personalized models are jointly learned for all users and how to incorporate feedback from the student models to learn more optimal teacher models?
Future work will explore the joint learning of the global model and optimal distillation parameters for each user, \ie joint optimization rather than the discrete formulation of \sys.
In this way, we plan to incorporate feedback from the student model after distillation to improve each user's optimal teacher model in the subsequent distillation steps. 

\shortsectionBf{An Evaluation Platform for Personalized Models.}
There exists no common evaluation platform to evaluate the performance of personalized FL models. 
Our study of various personalized learning schemes shows that the approaches use different datasets and data-splits, and report different evaluation metrics to demonstrate personalized models' effectiveness.
For instance, we observe that per-user accuracy often is not reported, or the equitable notion of fairness is not discussed.
In light of these observations, we plan to develop an evaluation framework for personalized FL methods. 
The evaluation framework will include different real and synthetic datasets and data splitting strategies, and enable future works to easily compare their personalized FL methods with existing approaches in convergence rate, local accuracy, and communication-efficiency.

%

\section{Conclusions}
We present \sys\footnote{\sys code is available at \url{https://tinyurl.com/hdh5zhxs}.},  a personalized FL algorithm , which addresses the statistical heterogeneity issue between different clients' data to improve the FL performance. 
\sys finds the optimal teacher model of each client during the FL training phase and distills the useful knowledge from optimal teachers into each user's local data after the training phase.
We evaluate the effectiveness of \sys on CIFAR-10 and MNIST datasets using three different data-splitting strategies. 
Experimentally, we show that \sys outperforms the \texttt{FedAvg} and three state-of-the-art personalized FL methods, \texttt{pFedMe}, \texttt{Per-FedAvg} and \texttt{FedPer} on the majority of data-splits with minimal communication cost. 
We additionally provide a set of numerical experiments to demonstrate the performance of \sys on different distillation objectives, how this performance is affected by the equitable notion of fairness among clients, and the number of communication rounds between clients and server.

\bibliographystyle{ACM-Reference-Format}
\bibliography{paper}

\newpage
\appendix

\section{Optimal Distillation Parameters}
\label{appendix-1}

\begin{figure*}[hb!]
\includegraphics[width=0.95\textwidth]{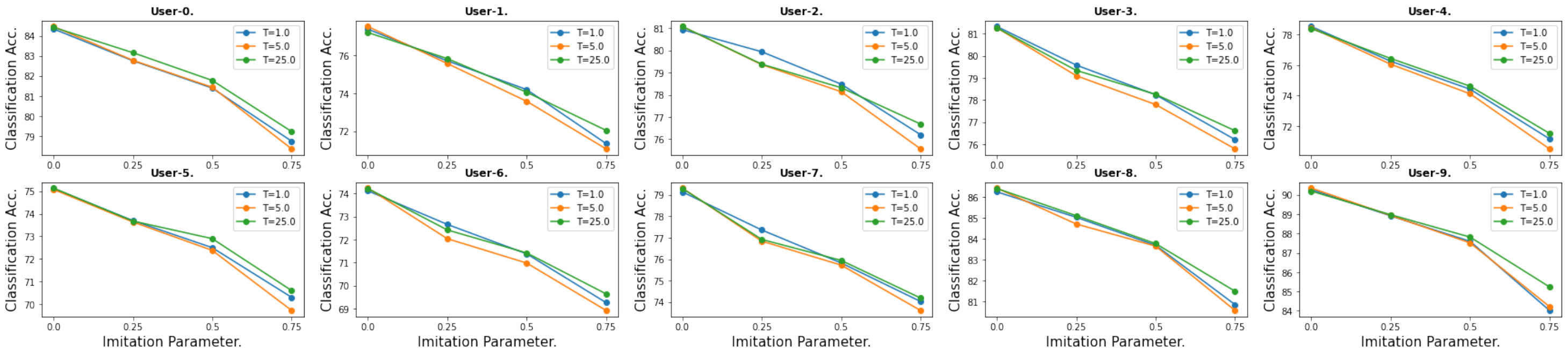}
\caption{Per-user interaction plots between $\mathtt{\lambda}$ and $\mathtt{T}$ on \textbf{DS-1} of CIFAR-10 averaged over the experimental runs.}
\label{fig:cifar10_ds1}
\end{figure*}

\begin{figure*}[hb!]
\begin{center}
\centerline{\includegraphics[width=0.95\textwidth]{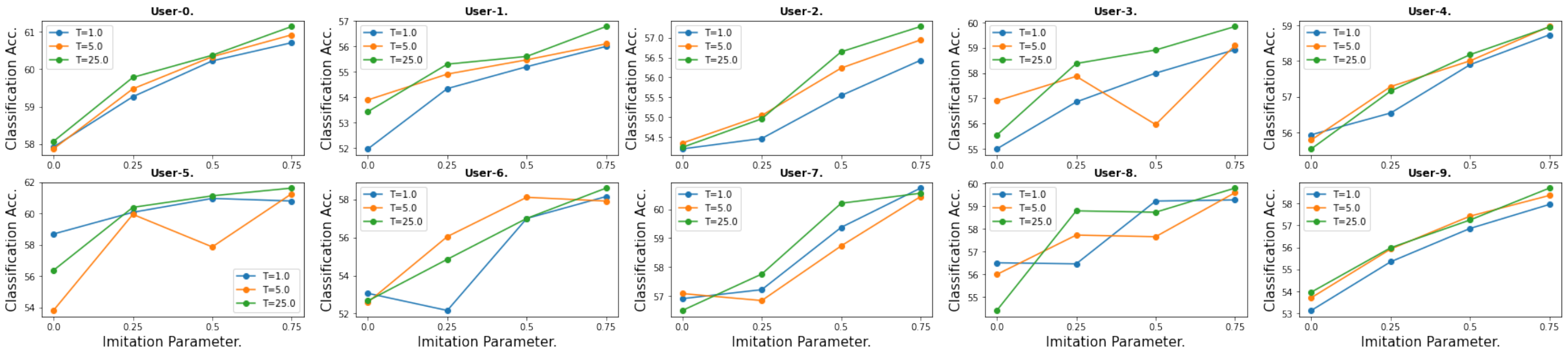}}
\caption{Per-user interaction plots between $\mathtt{\lambda}$ and $\mathtt{T}$ on \textbf{DS-2} of CIFAR-10 averaged over the experimental runs.}
\label{fig:cifar10_ds2}
\end{center}
\end{figure*}

\begin{figure*}[hb!]
\begin{center}
\centerline{\includegraphics[width=0.95\textwidth]{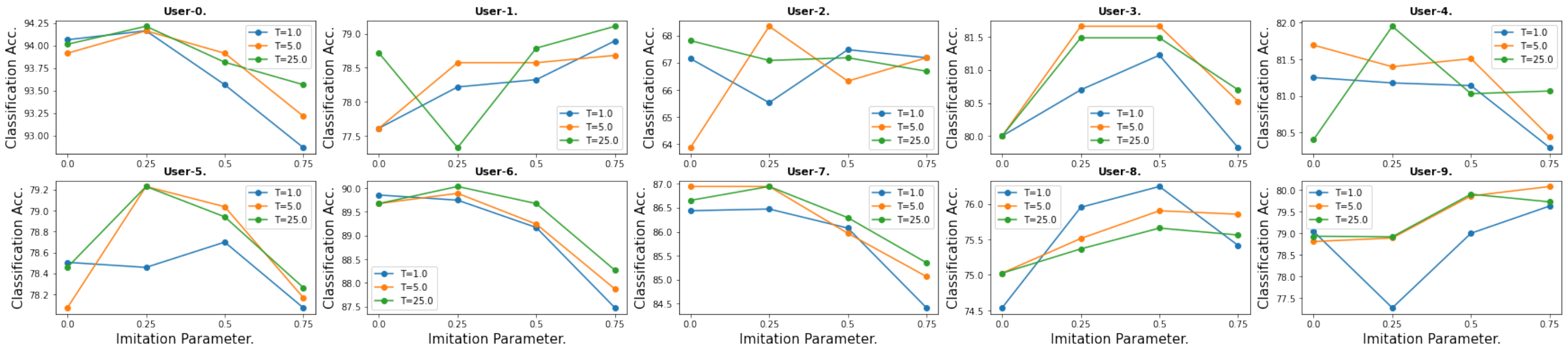}}
\caption{Per-user interaction plots between $\mathtt{\lambda}$ and $\mathtt{T}$ on \textbf{DS-3} of CIFAR-10 averaged over the experimental runs.}
\label{fig:cifar10_ds3}
\end{center}
\end{figure*}

\begin{figure*}[b!]
\begin{center}
\centerline{\includegraphics[width=0.95\textwidth]{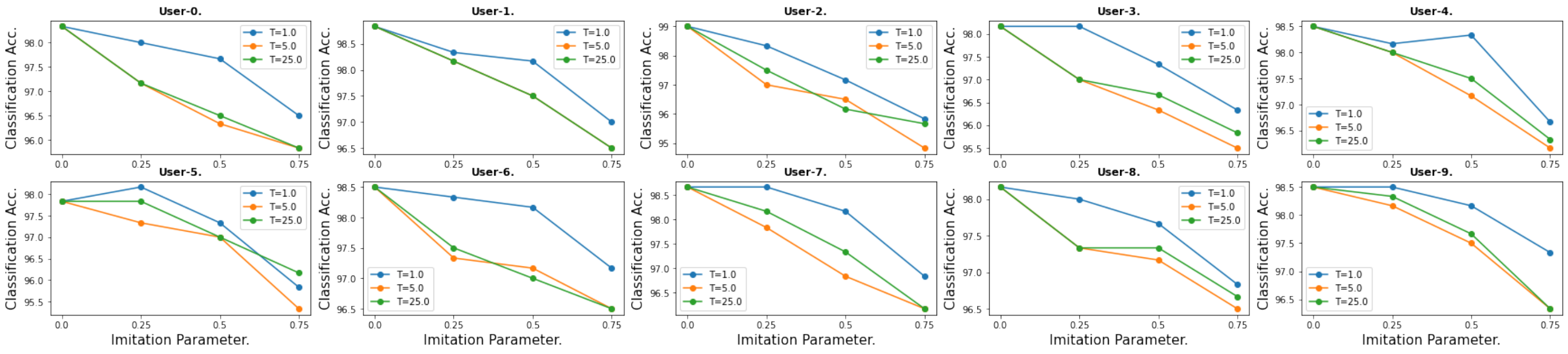}}
\caption{Per-user interaction plots between $\mathtt{\lambda}$ and $\mathtt{T}$ on \textbf{DS-1} of MNIST averaged over the experimental runs.}
\label{fig:mnist_ds1}
\end{center}
\end{figure*}

\begin{figure*}[hb!]
\begin{center}
\centerline{\includegraphics[width=0.95\textwidth]{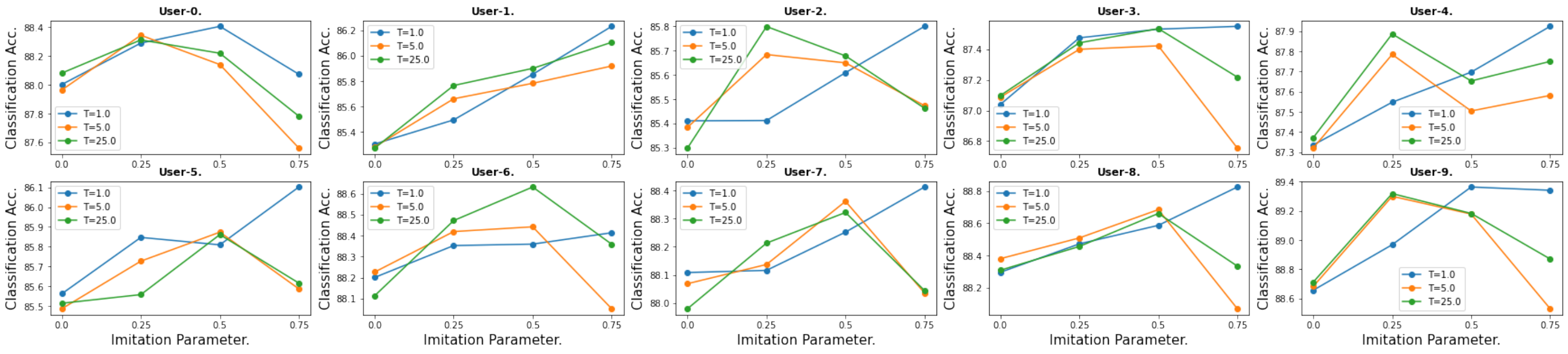}}
\caption{Per-user interaction plots between $\mathtt{\lambda}$ and $\mathtt{T}$ on \textbf{DS-2} of MNIST averaged over the experimental runs.}
\label{fig:mnist_ds2}
\end{center}
\end{figure*}

\begin{figure*}[hb!]
\begin{center}
\centerline{\includegraphics[width=0.95\textwidth]{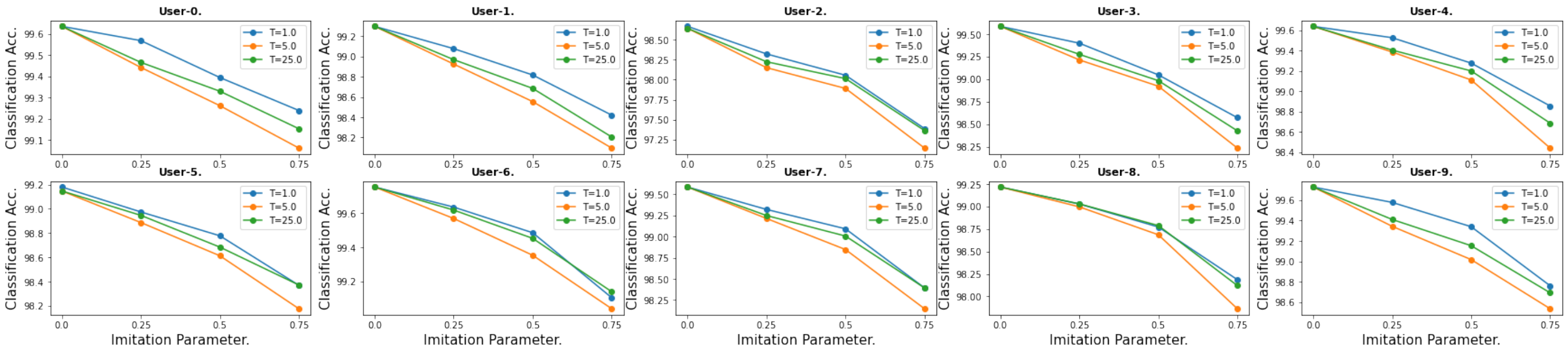}}
\caption{Per-user interaction plots between $\mathtt{\lambda}$ and $\mathtt{T}$ on \textbf{DS-3} of MNIST averaged over the experimental runs.}
\label{fig:mnist_ds3}
\end{center}
\end{figure*}

We present a detailed analysis of the selection of the optimal parameters introduced in~\ref{optParamSelection}. Figures~\ref{fig:cifar10_ds1}, \ref{fig:cifar10_ds2}, and \ref{fig:cifar10_ds3} show the classification accuracy on the test set of each user for a combination of values of $\mathtt{\lambda}$ and $\mathtt{T}$ averaged out over the five experimental runs on CIFAR-10. 
Figures~\ref{fig:mnist_ds1}, \ref{fig:mnist_ds2}, and \ref{fig:mnist_ds3} show the classification accuracy in the case of MNIST. The x-axes in these plots represent different values of the imitation parameters, and the y-axes represent the classification accuracy. We do not include the imitation parameter of $\mathtt{1}$ in these figures because it is never the case in our experiments that the optimal value of $\mathtt{\lambda}$ turns out to be $\mathtt{1}$.


\end{document}